\documentclass{article}

\usepackage{PRIMEarxiv}

\usepackage{amsmath,amsfonts}
\usepackage{algorithmic}
\usepackage{algorithm}
\usepackage{array}
\usepackage[caption=false,font=normalsize,labelfont=sf,textfont=sf]{subfig}
\usepackage{textcomp}
\usepackage{stfloats}
\usepackage{url}
\usepackage{verbatim}
\usepackage{graphicx}
\usepackage{cite}

\usepackage{colortbl}
\usepackage{xcolor}
\usepackage{color}
\usepackage{multirow}

\title{Facilitating Medical Imaging Translation via Dynamic Frequency Balance and Modality-Centered Knowledge Guidance
}

\author{%
  Dawei Zhou\\
  City University of Macau
  \And
  Jiahua Xu\\
  Xidian University
  \And
  Yizhong Zhao\\
  Communication University of China
  \And
  Nannan Wang\\
  Xidian University
  \And
  Cheng Lu\\
  Guangdong Provincial People's Hospital
  \And
  Zaiyi Liu\\
  Guangdong Provincial People's Hospital
  \And
  Lei Hu\\
  Guangdong Provincial People's Hospital
}

\begin{document}
\maketitle
\begin{abstract}
Multi-modal medical images play a crucial role in the precise and comprehensive clinical diagnosis. Diffusion model is a powerful strategy to synthesize the required medical images. However, existing approaches still suffer from the problem of \textbf{anatomical structure distortion} due to the overfitting of high-frequency information and the weakening of low-frequency information. Thus, we propose a novel method based on dynamic frequency balance and modality-centered knowledge guidance. Specifically, we first extract the low-frequency and high-frequency components by decomposing the critical features of the model using wavelet transform. Then, a \textbf{dynamic frequency balance} module is designed to adaptively adjust frequency for enhancing global low-frequency features and effective high-frequency details as well as suppressing high-frequency noise. To further overcome the challenges posed by the large differences between different medical modalities, we construct a \textbf{modality-centered knowledge-guided} mechanism that fuses the prior clinical modality-knowledge from a visual language model with visual features, to facilitate the generation of accurate anatomical structures. Experimental evaluations on multiple datasets show the proposed method achieves significant improvements in qualitative and quantitative assessments, verifying its effectiveness and superiority.
\end{abstract}
\section{Introduction}
\label{sec:intro}
Multi-modal images are crucial for comprehensive evaluation in medical image analysis, which can compensate for the limitations of a single modality and achieve a precise diagnosis \cite{liu2018digital, shen2019brain, tong2017multi}. In medical image analysis, especially brain tumor diagnosis, computed tomography (CT) and various magnetic resonance (MR) imaging techniques, including T1-weighted, T2-weighted, fluid-attenuated inversion recovery (FLAIR) and diffusion-weighted imaging (DWI), are widely used. Different modalities can provide different contrasts and complementary information for the clinical diagnosis \cite{kim2024adaptive, menze2014multimodal, shukla2017advanced, patel2017t2, mintorovitch1991comparison, moseley1990early}. However, in the clinical setting, multi-modal image acquisition is often limited by factors such as scanning costs and safety, resulting in missing data for some modalities. The lack of these key modalities can affect the accuracy of clinical diagnosis and treatment. These acquisition constraints motivate the development of cross-modal synthesis methods that estimate an unavailable modality from an acquired image \cite{shen2017deep, van1999multiple, sauerbrei1999building}. 

\begin{figure}[!t]
    \centering
    \includegraphics[width=6.2in]{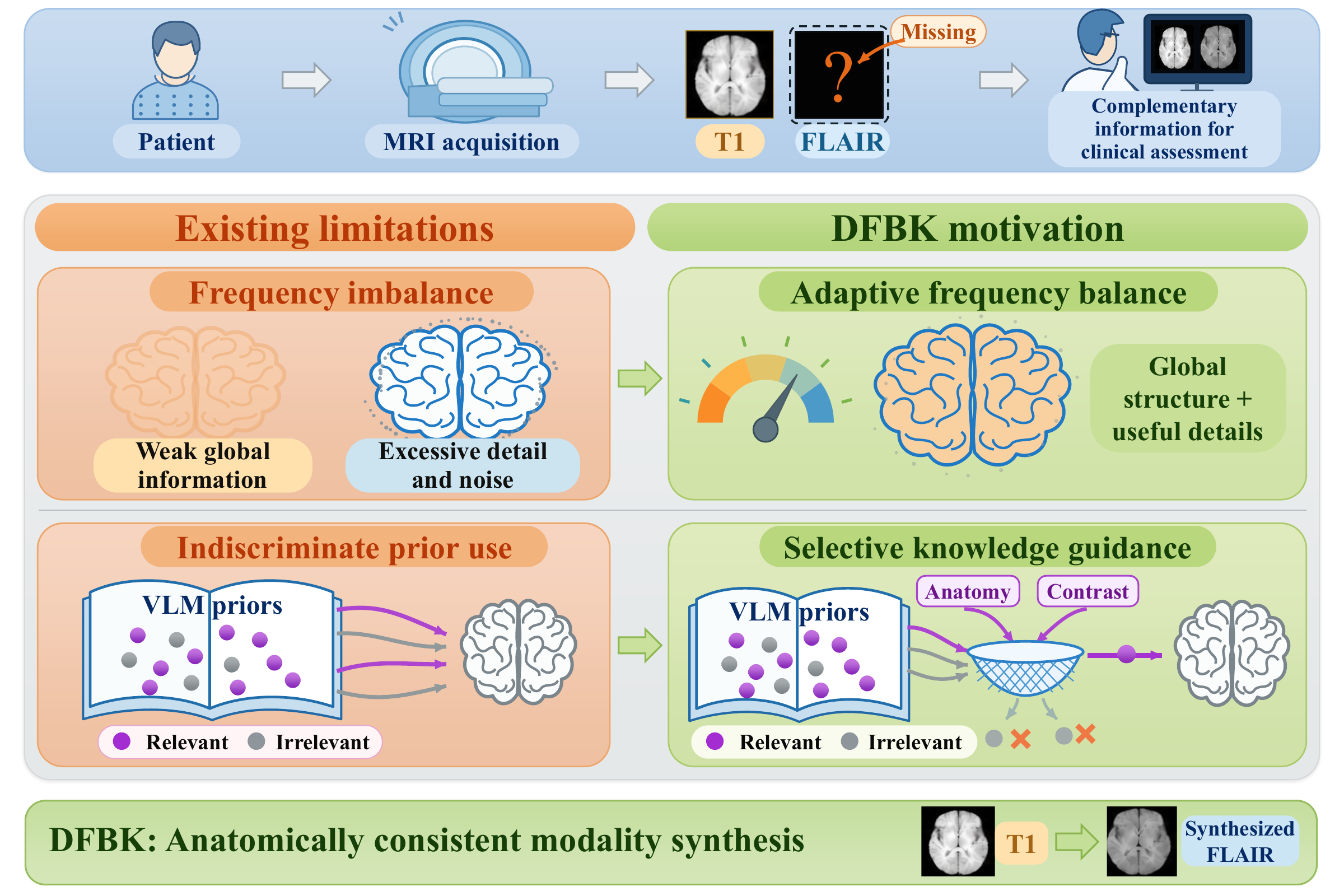}
    \caption{Motivation for DFBK. In clinical practice, certain imaging modalities may be unavailable due to acquisition costs and patient safety concerns, making it challenging for radiologists to make a comprehensive diagnosis. DFBK aims to address two challenges in modality synthesis: frequency imbalance and indiscriminate use of VLM priors. The dynamic frequency balance enhances global low-frequency features and informative high-frequency details while suppressing noise. The modality-centered knowledge guidance selectively incorporates relevant priors to support anatomy-consistent synthesis.}
    \label{motivation}
\end{figure}

Diffusion models formulate image generation through a forward noising process and a learned reverse denoising process \cite{song2020denoising, ho2020denoising}.This formulation has been applied to image-to-image translation \cite{choo2024slice, kim2024adaptive, nguyen2024ct, ozbey2023unsupervised}. It constructs a progressive generation framework by defining the forward process and the reverse process through rigorous mathematical derivation, with a more stable training process and a finer generation quality than GANs (Generative Adversarial Networks) \cite{dhariwal2021diffusion}. However, existing algorithms tend to overfit high-frequency information, making it difficult for the model to capture low-frequency features. This is caused by inconsistencies in signal intensity across identical anatomical structures within the same modality. Besides, the generated images are prone to suffer from \textit{anatomical structure distortion} due to the large differences between different medical modalities \cite{si2024freeu, xiao2024multi}.

Pretrained vision–language models (VLMs) offer an additional source of prior knowledge for guiding image translation \cite{korkmaz2024i2i, dalmaz2022resvit, li2024blip}. The pre-trained VLM based on the billion-parameter scale provides rich prior knowledge for image translation, greatly reducing the learning difficulty while ensuring that the generated images have the correct anatomical structure. However, existing algorithms often \textit{blindly} accept all prior knowledge, resulting in a failure to effectively fuse prior knowledge with the model, which has become a challenge in image translation \cite{xie2024sana}.

To address above issues, as shown in Fig. \ref{motivation}, we propose a novel translation method called DFBK, which is based on Dynamic Frequency Balance and modality-centered Knowledge guidance. Specifically, we first extract the low-frequency and high-frequency components by decomposing the critical features of the model using wavelet transform. Subsequently, we designed a dynamic frequency balance module to \textit{adaptively adjust frequency} for enhancing global low-frequency features and effective high-frequency details while suppressing high-frequency noise, thereby facilitating the model to generate correct anatomical structures. To address the challenges posed by the large differences between different medical modalities, we propose a novel modality-centered knowledge-guided mechanism. The mechanism \textit{fuse prior clinical knowledge with global low-frequency features} using pre-trained VLM. For high-frequency features, we only \textit{retain the critical prior clinical knowledge while suppressing irrelevant information to avoid the over-enhancement of high-frequency features}. This mechanism further achieves frequency balance while fusing prior knowledge and avoids the anatomical structure distortion caused by blindly accepting all prior knowledge. 

The main contributions are summarized as follows:
\begin{itemize}
    \item Diffusion model-based medical image translation can trigger anatomical structure distortion due to inappropriate utilization of frequency information, and we propose a novel method DFBK to mitigate this problem.
    \item We design a dynamic frequency balance module to enhance low-frequency and useful high-frequency components while suppressing high-frequency noise, thus adaptively and optimally leveraging frequency information.
    \item We construct a modality-centered knowledge-guided mechanism that fuses prior clinical knowledge with model features, which effectively improves the translation ability against large differences between different medical modalities.
    
    \item Our method is evaluated on multiple medical image translation tasks, which exhibits more accurate anatomical structures and superior quantitative metrics scores.
\end{itemize}

\section{Related Work}
\label{sec:work}

GAN-based approaches have been widely used in image translation \cite{armanious2019unsupervised, armanious2020unsupervised, chung2021simultaneous}. For example, Pix2pix \cite{isola2017image} achieves pixel-level synthesis of paired images through a simple network; CycleGAN \cite{zhu2017unpaired} introduces cyclic consistency loss to construct cross-domain image mapping. RevGAN \cite{van2019reversible} is further optimised by designing an approximately reversible architecture to maintain constant memory complexity. Ea-GAN \cite{yu2019ea} enhances the synthesis of 3D MR images by integrating edge information. ResViT \cite{dalmaz2022resvit} combines a Transformer architecture and fine-tuning techniques to capture contextual features. MaskGAN \cite{phan2023structure} introduces a mask generator to control the generated region for accurate structure mapping. In order to solve the problem of unstable quality of GAN generated images, SynDiff \cite{ozbey2023unsupervised} designed a loop-consistent architecture using a diffusion model to achieve bidirectional inter-modal translation. ResShift \cite{yue2024efficient} designed a diffusion model for direct translation, which reduces the number of diffusion steps and improves the translation efficiency. SelfRDB proposes a novel self-consistent recursive diffusion bridge to improve the performance of medical image transformation \cite{arslan2025self}. However, these approaches usually focus excessively on high-frequency features, resulting in inaccurate low-frequency anatomical structures. Therefore, a method is needed to balance the focus on different frequency information.

With the rapid development of VLM, text-driven approaches are gradually applied to image translation \cite{shiohara2024face2diffusion, fei2024dysen}. StyleCLIP \cite{patashnik2021styleclip} enables CLIP-guided feature manipulation by inverting the source image and obtaining its latent encoding. I2I-Galip \cite{korkmaz2024i2i}, on the other hand, uses pre-trained VLM, eliminating the need to individually train the generator-discriminator pairs for each pair of source-target images, thus improving the conversion efficiency and enhancing the accuracy of low-frequency anatomical structures. However, existing algorithms are crude in fusing prior knowledge and are unable to fuse effectively with models. Therefore, there is a need to explore a novel mechanism to effectively fuse prior knowledge to improve model performance.

\section{Methodology}

\subsection{Preliminary}
\noindent\textbf{Notice.}
We use lower-case letters $x_0$ and $y_0$ to denote the images of target modality and source modality, respectively. Moreover, we use $t$ to represent the timestep that the diffusion model is going through and $T$ to represent the total number of timesteps. We use the diffusion model based on ResShift and set $t \in \{1,2,3,4\}$. We use $\mathcal{N}(\cdot;\mu, \sigma^2I)$ to denote a variable that obeys a Gaussian distribution with mean $\mu$ and variance matrix $\sigma^2$.

\begin{figure*}[t]
    \centering
    \includegraphics[width= 6.2in]{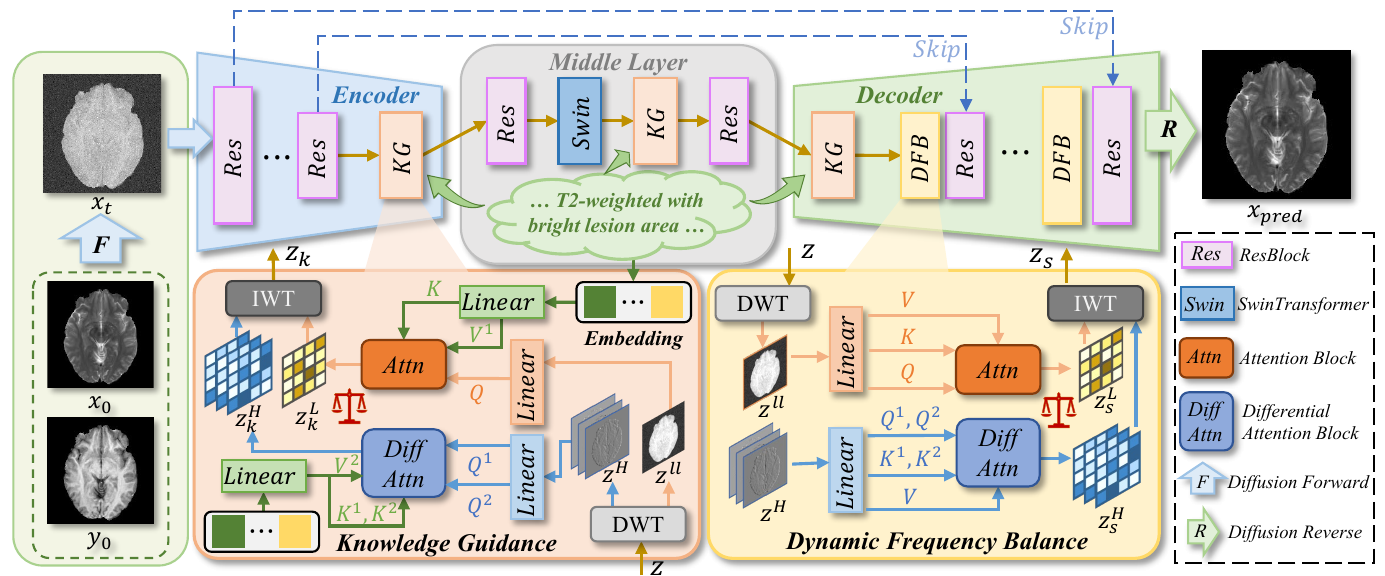}
    \caption{Framework of the proposed method DFBK. In the lower right corner is the Dynamic Frequency Balance module, which achieves frequency balance by adaptively enhancing low-frequency features and effective high-frequency features while suppressing high-frequency noise. In the lower left corner is the modality-centered Knowledge-Guided mechanism, which fuses the prior knowledge with global low-frequency features and retains only critical prior knowledge and suppresses irrelevant information when fusing with high-frequency features to avoid over-enhancing high-frequency features. Embedding is obtained from the text information using VLM's text encoder. By using the dynamic frequency balance module and the modality-centered knowledge-guided mechanism, we can generate target modal images with accurate anatomical structures.}
    \label{method}
\end{figure*}

\noindent\textbf{Discrete Wavelet Transformation (DWT).} DWT decomposes the image into low-frequency and high-frequency components. The low-frequency component represents the anatomical structure, while the high-frequency components represent the edges and contour details of the image. In this paper, we use Haar wavelet for DWT, which is widely used in applications \cite{phung2023wavelet, miao2024waveface}.

Given an image or a feature $z\in \mathbb{R}^{H\times W \times C}$, its low-frequency component $z^{ll}\in \mathbb{R}^{\frac{H}{2}\times \frac{W}{2} \times C}$ and high-frequency components $z^{lh}$, $z^{hl}$ and $z^{hh}\in \mathbb{R}^{\frac{H}{2}\times \frac{W}{2} \times C}$ can be decomposed by DWT,
\begin{equation}
\label{eq:1}
z^{ll},z^{lh},z^{hl},z^{hh}=DWT(z).
\end{equation}
where $z^{ll}$, $z^{lh}$, $z^{hl}$, and $z^{hh}$ denote the low--low (LL), low--high (LH), high--low (HL), and high--high (HH) subbands, respectively. 

DWT can not only effectively decompose frequency components, but also reduce the size of input images or features to reduce computational costs and speed up inference.

\noindent\textbf{ResShift.} 
ResShift uses a custom efficient diffusion process to achieve image translation. Specifically, given a pair of images $x_0$ and $y_0$, the forward process gradually adds noise to $x_0$ to obtain a Gaussian distribution with $y_0$ as the mean. This is represented as follows,
\begin{equation}
\label{eq:2}
q(x_t|x_{t-1}, y_0)=\mathcal{N}\left(x_t;x_{t-1}+\alpha_t e_0, \kappa^2\alpha_tI\right),
\end{equation}
where $e_0=y_0-x_0$, which represents the residual between the images of source modality and target modality. $\alpha_t=\eta_t-\eta_{t-1}$, where $\{\eta_t\}^T_{t=1}$ is a sequence that increases monotonically with timestep $t$ and satisfies $\eta_1 \to 0$ and $\eta_T \to 1$. $\kappa$ is a parameter controlling the variance. $I$ is the identity matrix.

Then, we can get the reverse process,
\begin{equation}
\label{eq:3}
q(x_{t-1}|x_t, y_0)=\mathcal{N}\left(x_{t-1};\mu_\theta(x_t,y_0,t), \kappa^2 \frac{\eta_{t-1}}{\eta_{t}}\alpha_t I\right),
\end{equation}
where $\mu_\theta(x_t,y_0,t)=\frac{\eta_{t-1}}{\eta_t}x_t+\frac{\alpha_t}{\eta_t}f_\theta(x_t, y_0, t)$ and $f_\theta$ is a neural network with parameter $\theta$ used to predict $x_0$.

To train the network $f_\theta$, we can obtain the objective function as follows,
\begin{equation}
\label{eq:4}
\mathop{\min}_{\theta} \sum_t{\frac{\alpha_t}{2\kappa^2\eta_t\eta_{t-1}}||f_\theta(x_t, y_0, t)-x_0||^2_2}.
\end{equation}

\subsection{Medical Image Translation with DFBK}
In this section we will introduce the detailed structure of DFBK. The framework is shown in Fig. \ref{method}. Firstly, we introduce how to build the dynamic frequency balance module. Then, we will introduce the principles and structure of the knowledge-guided mechanism. Finally, we combine them to obtain a network framework based on dynamic frequency balance and knowledge guidance.

\noindent\textbf{Dynamic Frequency Balance.} To address the problem that existing approaches overfit high-frequency information and weaken low-frequency information, which results in anatomical structure distortion, we proposed a dynamic frequency balance (DFB) module. As shown in Fig. \ref{method}, we first use Eq. \ref{eq:1} to decompose the features. Then, we use the self-attention mechanism \cite{vaswani2017attention} to enhance the global low-frequency features,
\begin{equation}
\label{eq:5}
z^{L}_s=S(\frac{Q_{z^{ll}}K_{z^{ll}}^T}{\sqrt{d}})V_{z^{ll}},
\end{equation}
where $S(\cdot)$ represents the $softmax$ function and $Q_{z^{ll}}$, $K_{z^{ll}}$, $V_{z^{ll}}$ represent the query, key and value obtained from $z^{ll}$ through a linear layer, respectively. $z_s^L$ is the enhanced low-frequency features and $d$ is the scaling factor.

In order to achieve dynamic frequency balance, we use the differential attention \cite{ye2024differential} mechanism for high-frequency features,
\begin{equation}
\label{eq:6}
z^{H}_s=\left(S(\frac{Q_{z^{H}}^1{K_{z^{H}}^1}^T}{\sqrt{d}})-\lambda \cdot S(\frac{Q_{z^{H}}^2 {K_{z^{H}}^2}^T}{\sqrt{d}})\right)V_{z^{H}},
\end{equation}
where $z^H \in \mathbb{R}^{H\times W \times 3C}$ is obtained by concatenating $z^{lh}$, $z^{hl}$ and $z^{hh}$ according to channel dimensions. $Q_{z^{H}}^1$, $Q_{z^{H}}^2$, $K_{z^{H}}^1$, $K_{z^{H}}^2$, and $V_{z^{H}}$ are all obtained from $z^{H}$ through a linear layer, where $Q$ and $K$ represent queries and keys, respectively, and $V$ represents the value. $\lambda$ is a learnable parameter. In order to synchronize the learning dynamics, we re-parameterize $\lambda$ as follows,
\begin{equation}
\label{eq:7}
\lambda=e^{(\lambda_q^1\cdot\lambda_k^1)}-e^{(\lambda_q^2\cdot\lambda_k^2)}+\lambda_{init},
\end{equation}
where $\lambda_q^1$, $\lambda_q^2$, $\lambda_k^1$, $\lambda_k^2$ are learnable parameters, and $\lambda_{init}$ is a constant, as defined in \cite{ye2024differential}. Differential attention can dynamically filter out and enhance effective high-frequency information while suppressing useless high-frequency noise. 

Then, we use the Inverse Wavelet Transformation (IWT) to recombine the frequency features,
\begin{equation}
\label{eq:8}
z_s=IWT(z^L_s,z^H_s),
\end{equation}
where $z_s$ is the recombination feature. The use of the dynamic frequency balance module can effectively strengthen the model's attention to low-frequency features, while dynamically adjusting the model's attention area for high-frequency features to enhance useful information and suppress high-frequency noise.

\noindent\textbf{Modality-centered Knowledge Guidance.} In order to effectively introduce VLM's prior clinical knowledge while ensuring the balance of frequency information, we designed a novel modality-centered knowledge-guided mechanism (KG) as shown in Fig. \ref{method}, which is similar to the dynamic frequency balance module. Specifically, we first fuse prior clinical knowledge and low-frequency features decomposed by Eq. \ref{eq:1},
\begin{equation}
\label{eq:9}
z^{L}_k=S(\frac{Q_{z^{ll}}K_{c}^T}{\sqrt{d}})V_{c}^1,
\end{equation}
where $K_{c}$ and $V_{c}^1$ represent the key and value obtained from the text embedding $c$ through a linear layer, and the text embedding is obtained from the text through the BiomedCLIP's \cite{zhang2023large} text encoder.

\begin{algorithm}[h]
    \footnotesize
    \caption{DFBK for medical image translation}
    \label{alg}
    \renewcommand{\algorithmicrequire}{\textbf{Input:}}
    \renewcommand{\algorithmicensure}{\textbf{Output:}}
    \renewcommand{\algorithmiccomment}[1]{\hfill $\triangleright$ #1}
    \begin{algorithmic}[1]
        \renewcommand{\baselinestretch}{1} 
        \selectfont
        \REQUIRE  $z$, $c$, $d$, $\lambda_q^1$, $\lambda_q^2$, $\lambda_k^1$, $\lambda_k^2$, $\lambda_{init}$.
        \ENSURE The recombination feature $z_s$ and $z_k$.
        
        \STATE $z^{ll},z^{lh},z^{hl},z^{hh} \gets DWT(z)$
        \STATE $z^H \gets concat(z^{lh},z^{hl},z^{hh})$
        \STATE $\lambda \gets e^{(\lambda_q^1\cdot\lambda_k^1)}-e^{(\lambda_q^2\cdot\lambda_k^2)}+\lambda_{init}$
        \IF {use \textit{DFB}}
        \STATE $Q_{z^{ll}},K_{z^{ll}},V_{z^{ll}} \gets Linear(z^{ll})$
        \STATE $z^{L}_s \gets S(\frac{Q_{z^{ll}}K_{z^{ll}}^T}{\sqrt{d}})V_{z^{ll}}$
        \STATE $Q_{z^{H}},K_{z^{H}},V_{z^{H}} \gets Linear(z^{H})$
        \STATE $z^{H}_s \gets (S(\frac{Q_{z^{H}}^1{K_{z^{H}}^1}^T}{\sqrt{d}})-\lambda \cdot S(\frac{Q_{z^{H}}^2 {K_{z^{H}}^2}^T}{\sqrt{d}}))V_{z^{H}}$
        \STATE $z_s \gets IWT(z^L_s,z^H_s)$ \COMMENT{Dynamic Frequency Balance}
        \ELSIF {use \textit{KG}}
        \STATE $Q_{z^{ll}} \gets Linear(z^{ll})$, $K_{c}, V_c^1 \gets Linear(c)$
        \STATE $z^{L}_k \gets S(\frac{Q_{z^{ll}}K_{c}^T}{\sqrt{d}})V_{c}^1$
        \STATE $Q_{z^{H}}^1, Q_{z^{H}}^2 \gets Linear(z^{H})$, $K_c^1, K_c^2, V_c^2 \gets Linear(c)$
        \STATE $z^{H}_k \gets (S(\frac{Q_{z^{H}}^1{K_{c}^1}^T}{\sqrt{d}})-\lambda \cdot S(\frac{Q_{z^{H}}^2 {K_{c}^2}^T}{\sqrt{d}}))V_{c}^2$
        \STATE $z_k \gets IWT(z^L_k,z^H_k)$ \COMMENT{Knowledge Guidance}
        \ELSE
        \STATE $z_s, z_k \gets z$
        \ENDIF
    
        \RETURN $z_s$ and $z_k$
        \renewcommand{\baselinestretch}{1} 
        \selectfont
    \end{algorithmic}
\end{algorithm}

Then, in order to avoid the over-enhancement of high-frequency features and ensure the frequency balance, we use differential cross attention to fuse high-frequency features with prior clinical knowledge,
\begin{equation}
\label{eq:10}
z^{H}_k=\left(S(\frac{Q_{z^{H}}^1{K_{c}^1}^T}{\sqrt{d}})-\lambda \cdot S(\frac{Q_{z^{H}}^2 {K_{c}^2}^T}{\sqrt{d}})\right)V_{c}^2,
\end{equation}
where $K_{c}^1$, $K_{c}^2$ and $V_{c}^2$ are all obtained from the text embedding through a linear layer.

Finally, we can obtain the recombination feature $z_k$,
\begin{equation}
\label{eq:11}
z_k=IWT(z^L_k,z^H_k).
\end{equation}

\noindent\textbf{Overall Framework.} We choose SwinUNet \cite{liu2021swin} as the base framework. Since the frequency balance in the backbone features is particularly important \cite{si2024freeu}, we use the frequency balance module on backbone features before skipping connection. In addition, introducing prior knowledge in all layers can lead to fusion mismatch, and introducing it in bottom layers is more effective \cite{wang2024instantstyle}, so we decide to use the knowledge-guided mechanism in bottom layers. The details of the overall process are presented in Algorithm \ref{alg}.

\section{Implementation Details}

\subsection{Experimental Datasets}
We utilize three public datasets for the evaluation and validation of DFBK. We use the Brain Tumor Segmentation challenge 2023 (BraTs 2023) \cite{bakas2017segmentation,bakas2017advancing,menze2014multimodal,baid2021rsna} dataset to perform common translation tasks such as T1 to T2 and T1 to FLAIR. In addition, we also perform experiments on rare translation tasks such as FLAIR to DWI and FLAIR to T1 using Ischemic Stroke Lesion Segmentation challenge 2015 (ISLES 2015) \cite{maier2017isles,maier2015extra}. To further verify the generalization ability, we also perform experiments on complex translation tasks of CT and MR using Synthesizing computed tomography for Radiotherapy 2023 (SynthRAD 2023) \cite{thummerer2023synthrad2023}. For BraTs dataset, we select 40,032 images from 1,251 subjects with 32 images per subject for training. Additionally, 7,008 images were chosen from 219 subjects with 32 images per subject for generating the target modality and evalution. For ISLES dataset, we select 1,792 images from 28 subjects with 64 images per subject for training and 2,304 images from 36 subjects with 64 images per subject for evalution. For SynthRAD dataset, we select 9,216 images from 144 subjects with 64 images per subject for training and 2,304 images from 36 subjects with 64 images per subject for evaluation.

\begin{figure*}[h]
    \centering
    \includegraphics[width= 6.2in]{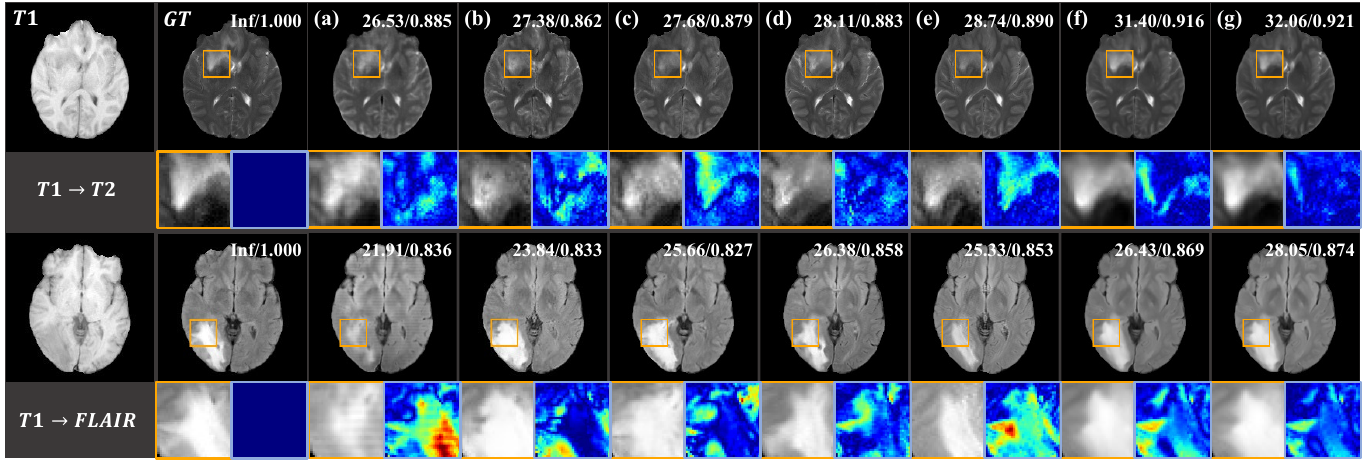}
    \caption{Comparison of approaches on BraTs: (a) Ea-GAN, (b) RevGAN, (c) MaskGAN, (d) ResViT, (e) SynDiff, (f) SelfRDB, (g) DFBK (ours). The first and third rows show the results of T1 to T2 and T1 to FLAIR translation, respectively. The first column is the source image, and the second column is the target ground-truth. The second and fourth rows show the zoomed-in anatomical structures for the first and third rows, respectively. PSNR and SSIM values of each image are shown in the corner of images. The yellow box indicates the zoomed-in visualization area, and the blue box represents the difference heatmap between the generated image and the ground truth. The color indicates the degree of difference from small to large, with the brighter color (e.g., red) reflecting the larger difference and vice versa (e.g., blue).}
    \label{brats}
\end{figure*}

\subsection{Comparison Approaches}
We select three classes of approaches for comparison. The first is GANs-based approaches, EaGAN, RevGAN and MaskGAN. The second is transformer-based approach, ResViT, which is a large model fine-tuning approach. The third is the diffusion-based approaches, SynDiff and SelfRDB. The training parameters are set according to the parameters specified in the original paper, respectively. We use two evaluation metrics, peak signal-to-noise ratio (PSNR) and structural similarity index measure (SSIM) to quantitatively evaluate the generated results to verify the effectiveness of our method.
\section{Experimental Results}\label{results}
\begin{figure*}[t]
    \centering
    \includegraphics[width= 6.2in]{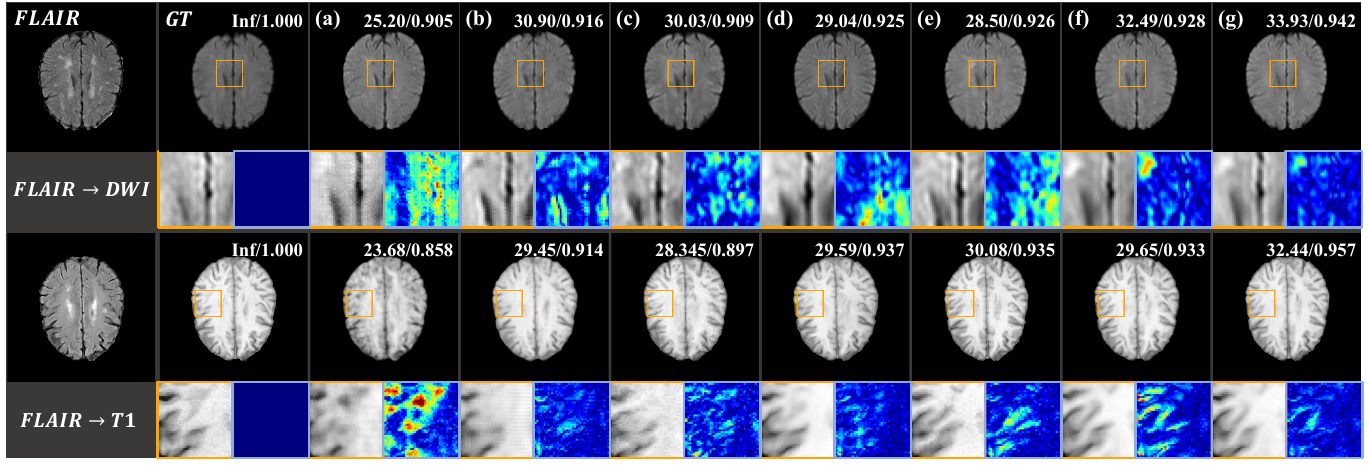}
    \caption{Comparison of approaches on ISLES: (a) Ea-GAN, (b) RevGAN, (c) MaskGAN, (d) ResViT, (e) SynDiff, (f) SelfRDB, (g) DFBK (ours). The first and third rows show the results of FLAIR to DWI and FLAIR to T1 translation, respectively. The first column is the source image, and the second column is the target ground-truth. The second and fourth rows show the zoomed-in anatomical structures for the first and third rows, respectively. PSNR and SSIM values of each image are shown in the corner of images.} 
    \label{SISS}
\end{figure*}

\begin{figure*}[t]
    \centering
    \includegraphics[width= 6.2in]{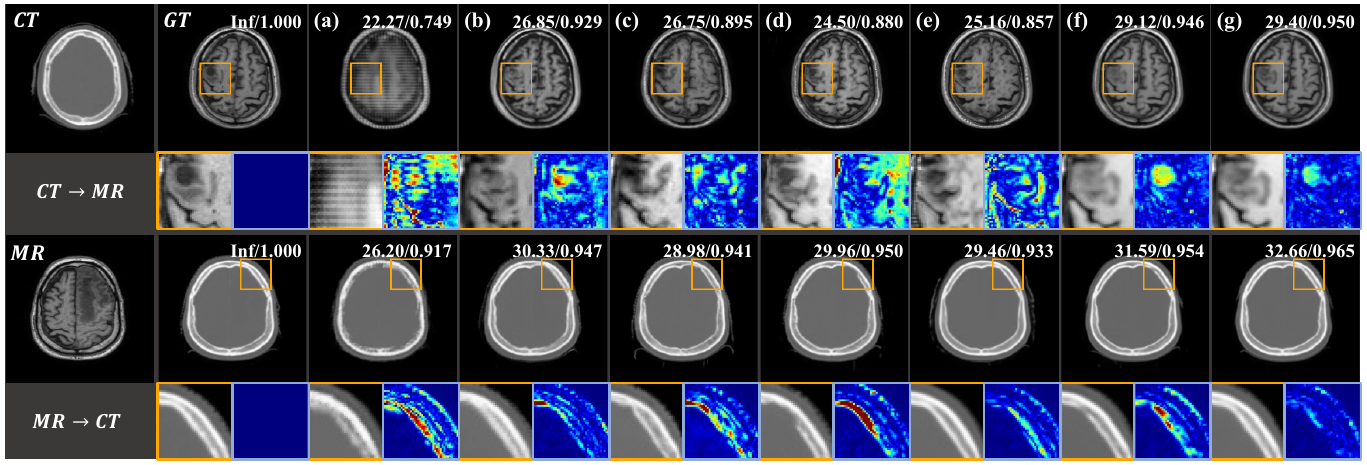}
    \caption{Comparison of approaches on SynthRAD: (a) Ea-GAN, (b) RevGAN, (c) MaskGAN, (d) ResViT, (e) SynDiff, (f) SelfRDB, (g) DFBK (ours). The first and third rows show the results of CT to MR and MR to CT translation, respectively. The first column is the source image, and the second column is the target ground-truth. The second and fourth rows show the zoomed-in anatomical structures for the first and third rows, respectively. PSNR and SSIM values of each image are shown in the corner of images.}
    \label{syn}
\end{figure*}

\subsection{Qualitative Results}

Fig. \ref{brats} shows the results of various methods for medical image translation on BraTs. Ea-GAN introduces grating-like artifacts, and the anatomical structure in some areas is very blurred or even completely distorted as shown in Fig. \ref{brats} (a). RevGAN has unrecognizable anatomical structures and introduces additional lesion areas (Fig. \ref{brats} (b)). MaskGAN suffers from generating incorrect content and the synthesized images appearing overly blurred (Fig. \ref{brats} (c)). ResViT generates a relatively correct anatomical structure, but the brightness of some areas does not match the target modality (Fig. \ref{brats} (d)). The anatomical structure generated by SynDiff is incorrect, and the brightness do not match the target modality (Fig. \ref{brats} (e)). SelfRDB still retains some information from the original modality, leading to inaccurate anatomical structure (Fig. \ref{brats} (f)). In contrast, the images generated by our proposed method demonstrate significantly improved anatomical accuracy and more precise contrast representation. As shown in Fig. \ref{brats} (g), our approach effectively preserves the structural integrity of anatomical features while maintaining the correct brightness distribution according to target modality. Furthermore, our method achieves the better performance across the quantitative evaluation metrics, underscoring its superiority in medical image translation.

Fig. \ref{SISS} shows the results of various methods for medical image translation on ISLES. Ea-GAN cannot generate the correct anatomical structure, and the content is completely distorted, as shown in Fig. \ref{SISS} (a). RevGAN introduces faint raster-like artifacts, and some areas are distorted (Fig. \ref{SISS} (b)). MaskGAN generates blurry content (Fig. \ref{SISS} (c)). The anatomical structure generated by ResViT (Fig. \ref{SISS} (d)) and SynDiff (Fig. \ref{SISS} (e)) exhibits noticeable incorrect, with significant distortions in shape and structure. SelfRDB has some unclear boundaries (Fig. \ref{SISS} (f)). In contrast, as shown in Fig. \ref{SISS} (g), our method presents more accurate anatomical structures and performs better.

We also shows the results on the difficult task of CT to MR translation on SynthRAD, as shown in Fig. \ref{syn}. Since CT images do not provide information about the internal anatomical structure of the brain, most approaches cannot generate correct or relevant target modality anatomical structures. As shown in Fig. \ref{syn} (a), Ea-GAN produces MR images with raster artifacts and lacks clear internal anatomical structures, while its CT images show blurred cranial boundaries. RevGAN can generate relevant MR structures but fails to delineate the lesion area, and its CT images also have blurred cranial boundaries (Fig. \ref{syn} (b)). MaskGAN confuses the lesion with normal brain structures in the MR images, and its CT images exhibit incorrect cranial boundaries (Fig. \ref{syn} (c)). Although ResViT (Fig. \ref{syn} (d)) and SynDiff (Fig. \ref{syn} (e)) localize the lesion area in generated MR images, they do not accurately reproduce the overall internal anatomy and CT images generated by ResViT miss part of the cranial boundaries. SelfRDB generates incorrect textures (Fig. \ref{syn} (f)). In contrast, our method accurately localizes the MR lesion area, generates correct anatomical structures, and produces CT images with correct cranial boundaries (Fig. \ref{syn} (g)).

\begin{table*}[t]
\setlength{\tabcolsep}{2.5pt}
\renewcommand\arraystretch{1.1}
\centering
\begin{small}
\caption{Quantitative comparison of different approaches on BraTs and ISLES. The values in the table represent the average values for the entire test dataset. \textbf{Bold} represents the best result and \underline{underlined} represents the suboptimal result.} 
\begin{tabular}{l|cccccc|cccccc}
\hline
Dataset & \multicolumn{6}{c|}{BRATS} & \multicolumn{6}{c}{ISLES} \\ 
Tasks & \multicolumn{3}{c}{T1--\textgreater{}T2} & \multicolumn{3}{c|}{T2--\textgreater{}T1} & \multicolumn{3}{c}{T1--\textgreater{}FLAIR} & \multicolumn{3}{c}{FLAIR--\textgreater{}T1} \\ \hline \hline
Method             & PSNR $\uparrow$         & SSIM $\uparrow$       & RMSE $\downarrow$       & PSNR $\uparrow$         & SSIM $\uparrow$         & RMSE $\downarrow$        & PSNR $\uparrow$         & SSIM $\uparrow$         & RMSE $\downarrow$         & PSNR $\uparrow$         & SSIM $\uparrow$         & RMSE $\downarrow$\\ \hline
Ea-GAN              & 22.43        & \underline{0.825}       & 0.083       & 21.16         & 0.773         & 0.091        & 24.18         & 0.850         & 0.065        & 21.35         & 0.810         & 0.092       \\
RevGAN             & 22.61        & 0.816       & \underline{0.080}       & \underline{21.83}         & 0.769         & \underline{0.085}        & 25.40         & 0.862         & 0.056        & 23.64         & 0.844         & 0.072\\
MaskGAN            & 21.40        & 0.809       & 0.092       & 20.41         & 0.735         & 0.101        & 25.24         & 0.856         & 0.057        & 23.15         & 0.834         & 0.077\\
ResViT             & \underline{22.89}        & 0.822       & \underline{0.080}       & 21.87         & \underline{0.786}         & \underline{0.085}        & \underline{25.67}         & 0.872         &\underline{0.055}        & \underline{24.90}         & \underline{0.873}         & \underline{0.064}\\
SynDiff            & 21.68        & 0.814       & 0.091       & 20.41         & 0.755         & 0.100        & 25.39         & 0.872         & 0.056        & 24.32         & 0.867         & 0.069\\
SelfRDB 
& 22.01 & 0.821 & 0.088  & 20.74 & 0.767 & 0.097 
& 25.27 & \underline{0.877} & 0.058 & 24.12 & 0.869 & 0.074
\\
\rowcolor[HTML]{E8E5E5}
DFBK (ours)               & \textbf{23.61}             & \textbf{0.846}            & \textbf{0.074}            & \textbf{22.38}              & \textbf{0.797}              & \textbf{0.080}             & \textbf{25.76}              & \textbf{0.879}              & \textbf{0.054}             & \textbf{25.42}              & \textbf{0.879}              & \textbf{0.060}            \\ \hline
\end{tabular}
\label{table:1}
\end{small}
\end{table*}

\begin{table}[t]
\setlength{\tabcolsep}{1.0pt}
\renewcommand\arraystretch{1.1}
\centering
\begin{small}
\caption{Quantitative comparison of different approaches on SynthRAD. The values in the table represent the average values for the entire test dataset. \textbf{Bold} represents the best result and \underline{underlined} represents the suboptimal result.} 
\begin{tabular}{l|ccc|ccc}
\hline 
Tasks & \multicolumn{3}{c|}{MR--\textgreater{}CT}                                      & \multicolumn{3}{c}{CT--\textgreater{}MR}                   \\ 
\hline \hline
Method                   & PSNR $\uparrow$          & SSIM $\uparrow$          & RMSE $\downarrow$           & PSNR $\uparrow$          & SSIM $\uparrow$          & RMSE $\downarrow$          \\ \hline
Ea-GAN                   & 24.75          & 0.852          & 0.059          & 20.39          & 0.679          & 0.098          \\
RevGAN                   & 26.24          & 0.872          & 0.051          & 22.94          & 0.798          & 0.073          \\
MaskGAN                  & 25.56          & 0.871          & 0.055          & 22.04          & 0.761          & 0.081          \\
ResViT                   & 26.92          & 0.887          & \underline{0.047}          & 22.88          & 0.801          & 0.075          \\
SynDiff                  & 26.05          & 0.877          & 0.052         & 20.73          & 0.705          & 0.094          \\
SelfRDB 
& \underline{27.06} & \underline{0.896} & \underline{0.047} & \underline{23.17} & \underline{0.823} & \underline{0.072} \\

\rowcolor[HTML]{E8E5E5}
DFBK (ours)              & \textbf{27.51} & \textbf{0.897} & \textbf{0.044} & \textbf{23.54} & \textbf{0.828} & \textbf{0.069} \\ \hline
\end{tabular}
\label{table:2}
\end{small}
\end{table}

\subsection{Quantitative Results}
Tab. \ref{table:1} and Tab. \ref{table:2} shows the mean values for each quantitative metric, where higher PSNR and SSIM values perform better. Our method outperforms all other approaches in six translation tasks on three datasets. Its superior performance stems from effectively balancing the model's extraction of high- and low-frequency information and effective prior knowledge guidance, thereby preventing anatomical distortions caused by too fast high-frequency convergence and insufficient low-frequency extraction. In contrast, GAN-based approaches produce poor results due to their inherent instability in training, which makes them particularly susceptible to outliers in low-quality datasets such as BraTs. SynDiff, which incorporates GAN principles into a diffusion model, also shows unsatisfactory performance because its more complex training process is more sensitive to data quality. ResViT achieves relatively good results by leveraging large model fine-tuning, but still causes anatomical distortion. In contrast, our method effectively achieves frequency balance and has better performance.

\begin{figure*}[t]
    \centering
    \includegraphics[width= 6.2in]{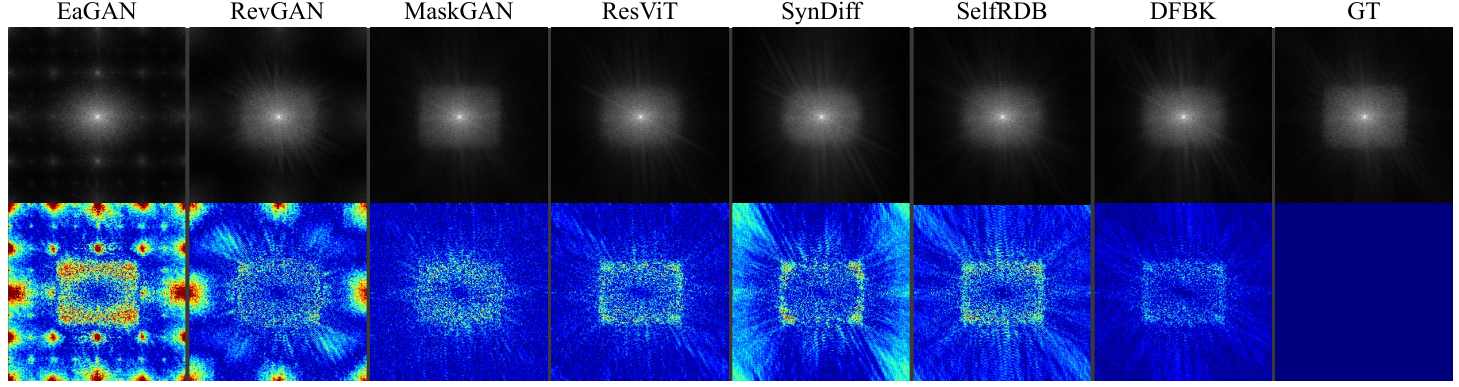}
    \caption{Comparison of approaches in the frequency spectrum. This experiment was conducted on the FLAIR--\textgreater{}DWI task of the ISLES dataset. The first row shows the average spectrogram of results produced by each approach, while the second row shows the difference heatmap between each approach’s spectrogram and the GT spectrogram. The color indicates the degree of difference from small to large, with the brighter color (e.g., red) reflecting the larger difference and vice versa  (e.g., blue).}
    \label{frequency_compare}
\end{figure*}

\begin{figure*}[t]
    \centering
    \includegraphics[width= 6.2in]{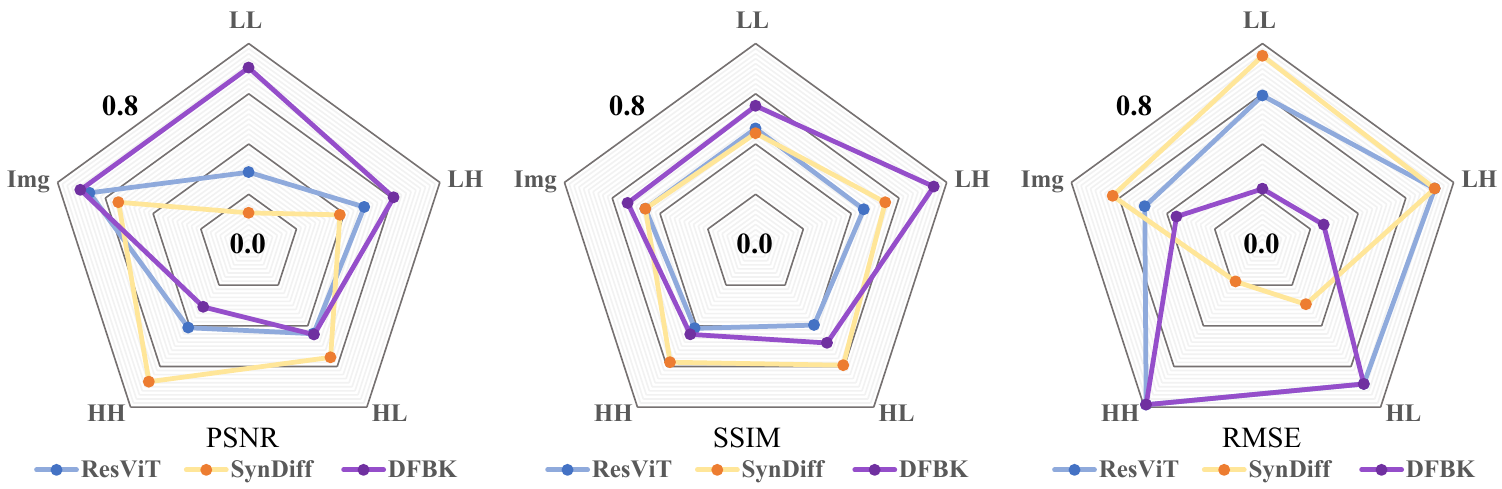}
    \caption{Comparison of radar charts on different approaches. This experiment was conducted on the FLAIR--\textgreater{}DWI task of the ISLES dataset. We selected ResViT and SynDiff, which demonstrated relatively strong performance, as comparative approaches. “Img” represents the overall image metrics reported in Tab.~\ref{table:1}. To facilitate intuitive comparison, all metrics were normalized. Specifically, for the metrics (PSNR, SSIM, RMSE) of each subband and the overall image, the maximum and minimum values among the three methods were first determined. The lower bound was set to 0.95 times the minimum value, and the upper bound to 1.05 times the maximum value, followed by min–max normalization to enhance the contrast and distinction of the radar chart. The results show that our method significantly improves the reconstruction quality in the low-frequency sub-band (LL), while selectively enhancing informative high-frequency details (LH) and suppressing high-frequency noise (HL, HH), thereby achieving a better balance between overall image quality, structural fidelity, and detail preservation (Img).}
    \label{wave}
\end{figure*}

\begin{table*}[t]
\setlength{\tabcolsep}{2.5pt}
\renewcommand\arraystretch{1.1}
\centering
\caption{Quantitative comparison of frequency components (LL, LH, HL, HH) across different methods. This experiment was conducted on the FLAIR--\textgreater{}DWI task of the ISLES dataset.}
\begin{small}
\begin{tabular}{l|ccc|ccc|ccc|ccc}
\hline
\multirow{2}{*}{Method} & \multicolumn{3}{c|}{LL} & \multicolumn{3}{c|}{LH} & \multicolumn{3}{c|}{HL} & \multicolumn{3}{c}{HH} \\ 
\cline{2-13} 
 & PSNR $\uparrow$ & SSIM $\uparrow$ & RMSE $\downarrow$ 
 & PSNR $\uparrow$ & SSIM $\uparrow$ & RMSE $\downarrow$ 
 & PSNR $\uparrow$ & SSIM $\uparrow$ & RMSE $\downarrow$ 
 & PSNR $\uparrow$ & SSIM $\uparrow$ & RMSE $\downarrow$ \\
\hline
EaGAN  & 24.14 & 0.865 & 0.100 & 28.56 & 0.662 & 0.015 & 24.91 & 0.627 & 0.018 & 28.72 & 0.634 & 0.010 \\
RevGAN & 24.95 & 0.886 & 0.091 & 29.50 & 0.725 & 0.014 & 25.08 & 0.671 & 0.017 & 30.91 & 0.688 & 0.008 \\
MaskGAN & 24.59 & 0.873 & 0.094 & 30.60 & 0.773 & 0.012 & 25.53 & 0.692 & 0.016 & 31.68 & 0.802 & 0.007 \\
ResViT & 24.91 & 0.891 & 0.091 & 30.87 & 0.801 & 0.012 & 25.84 & 0.719 & 0.016 & 31.92 & 0.840 & 0.007 \\
SynDiff & 24.63 & 0.889 & 0.094 & 30.46 & 0.809 & 0.012 & \textbf{26.18} & \textbf{0.737} & \textbf{0.015} & \textbf{33.28} & \textbf{0.857} & \textbf{0.006} \\
SelfRDB & 24.61 & 0.878 & 0.094 & 30.61 & 0.790 & 0.012 & 25.30 & 0.694 & 0.017 & 32.61 & 0.844 & \textbf{0.006} \\
\rowcolor[HTML]{E8E5E5}
DFBK (ours) & \textbf{25.63} & \textbf{0.900} & \textbf{0.084} & \textbf{31.36} & \textbf{0.827} & \textbf{0.011} & 25.85 & 0.727 & 0.016 & 31.40 & 0.843 & 0.007 \\
\hline
\end{tabular}
\label{tab:wavelet_metrics}
\end{small}
\end{table*}

\subsection{Frequency Component Analysis}
Due to differences in acquisition conditions, imaging parameters, and imaging devices, identical anatomical structures within the same modality exhibit significant differences in signal intensity distributions in the dataset. Such differences interfere with the model’s ability to effectively extract low-frequency anatomical features, resulting in generated images whose low-frequency anatomical structures deviate from the true morphology or are inaccurate, as shown in Fig. \ref{frequency_compare}.
As shown in Fig.~\ref{wave} and Tab.~\ref{tab:wavelet_metrics}, existing methods generally exhibit insufficient capability in generating low-frequency components, while suffering from overfitting in high-frequency subbands, which introduces redundant information and interferes with overall structural reconstruction. In contrast, our method effectively enhances the quality of low-frequency component generation, yielding more accurate and coherent structural representations. Meanwhile, it selectively promotes the generation of useful high-frequency details and suppresses noise in HL and HH subbands. This balance between low-frequency enhancement and high-frequency regulation contributes to clearer anatomical structures and improved overall reconstruction fidelity.

\subsection{Ablation Study}
In this section, we validate the effectiveness of dynamic frequency balance module and knowledge-guided mechanism. All experiments are performed on ISLES dataset.

\noindent\textbf{DFB and KG.}
We conduct ablation experiments on the dynamic frequency balance module and the knowledge-guided mechanism to verify their important roles in medical image translation. From Tab. \ref{table:3}, we can see that both dynamic frequency balance and knowledge guidance play an important role in generating the correct anatomical structure and affect most indicators, such as PSNR and SSIM. At the same time, it can be seen that when the source modality and target modality are significantly different, such as the conversion from FLAIR to T1, the knowledge-guided mechanism plays an important role and can effectively guide the model to generate correct anatomical structures.

\begin{table}[t]
\centering
\begin{small}
\setlength{\tabcolsep}{1.0pt}
\renewcommand\arraystretch{1.1}
\caption{Ablation study on Dynamic Frequency Balance (DFB) and Knowledge Guidance (KG). ``None" indicates the performances in the absence of DFB and KG.}
\begin{tabular}{l|ccc|ccc}
\hline
\multicolumn{1}{c|}{Task} & \multicolumn{3}{c|}{FLAIR--\textgreater{}DWI} & \multicolumn{3}{c}{FLAIR--\textgreater{}T1} \\ \hline \hline
Method                    & PSNR $\uparrow$          & SSIM $\uparrow$          & RMSE $\downarrow$          & PSNR $\uparrow$           & SSIM $\uparrow$         & RMSE $\downarrow$        \\ \hline
None                      & 25.61         & 0.877         & 0.055         & 25.15         & 0.876        & 0.064        \\
+DFB                      & 25.70         & 0.878         & 0.054         & 25.21         & 0.877        & 0.062        \\
+KG                       & 25.68         & 0.877         & 0.054         & 25.31         & 0.878        & 0.061        \\
+DFB, KG                   & \textbf{25.76}         & \textbf{0.879 }        & \textbf{0.054}         & \textbf{25.42}         & \textbf{0.879}        & \textbf{0.060}        \\ \hline
\end{tabular}
\label{table:3}
\end{small}
\end{table}

\noindent\textbf{The components in DFB.}
It can be seen from Tab. \ref{table:4} that only enhancing the low-frequency features (L) can alleviate the poor performance of the model due to too fast high-frequency convergence to a certain extent, while only dynamically balancing the high-frequency features (H) suppresses some high-frequency noise, but the enhancement of the effective high-frequency components actually accelerates the high-frequency convergence of the model and reduces the performance. The improvement of PSNR in the FLAIR to DWI task is related to its similar modality. To further demonstrate the importance of frequency balance, we replace the dynamic frequency balance module with a normal self-attention mechanism (NormAttn). The results show that the performance of the model has slightly decreased. This is because the self-attention mechanism can only enhance global features, but cannot balance high-frequency and low-frequency features, which makes the model further strengthen the convergence of high-frequency. In contrast, the dynamic frequency balance module can decompose and process frequency components more carefully.

\noindent\textbf{The components in KG.}
As shown in Tab. \ref{table:5}, only fusing the prior knowledge with low-frequency features can guide the model to enhance the extraction of low-frequency features and slightly improve the model performance. However, only fusing the prior knowledge with high-frequency features cannot achieve frequency balance, even if the enhancement of high-frequency features is suppressed. This shows that the enhancement of low-frequency is more important. To further verify the effectiveness of the knowledge-guided mechanism, we replaced it with a cross-attention mechanism (CrossAttn). The results show that the cross-attention mechanism can slightly improve the model performance, but it limits the guidance of the model by the prior knowledge due to the over-enhancement of high-frequency. In contrast, the knowledge-guided mechanism processes high and low frequencies separately, effectively improving model performance.

\begin{table}[t]
\centering
\begin{small}
\setlength{\tabcolsep}{1.0pt}
\renewcommand\arraystretch{1.1}
\caption{Ablation study on the components in the Dynamic Frequency Balance module. ``None" indicates the performances when there are no additional operations in this module.}
\begin{tabular}{l|ccc|ccc}
\hline
\multicolumn{1}{c|}{Task} & \multicolumn{3}{c|}{FLAIR--\textgreater{}DWI} & \multicolumn{3}{c}{FLAIR--\textgreater{}T1} \\ \hline \hline
Method                    & PSNR $\uparrow$          & SSIM $\uparrow$          & RMSE $\downarrow$          & PSNR $\uparrow$           & SSIM $\uparrow$         & RMSE $\downarrow$           \\ \hline
None                      & 25.61          & 0.877          & 0.055          & 25.15          & 0.876          & 0.064          \\
+L                        & 25.65          & 0.878          & 0.055          & 25.17          & 0.876          & 0.063          \\
+H                        & 25.67          & 0.877          & 0.055          & 25.12          & 0.874          & 0.063          \\
+NormAttn                 & 25.59          & 0.878          & 0.055          & 25.09          & 0.874          & 0.063          \\
+L, H                      & \textbf{25.70} & \textbf{0.878} & \textbf{0.054} & \textbf{25.21} & \textbf{0.877} & \textbf{0.062} \\ \hline
\end{tabular}
\label{table:4}
\end{small}
\end{table}

\begin{table}[t]
\centering
\begin{small}
\setlength{\tabcolsep}{1.0pt}
\renewcommand\arraystretch{1.1}
\caption{Ablation study on the components in the Knowledge-Guided mechanism. ``None" indicates the performances when there are no additional operations in this mechanism.}
\label{table:5}
\begin{tabular}{l|ccc|ccc}
\hline
\multicolumn{1}{c|}{Task} & \multicolumn{3}{c|}{FLAIR--\textgreater{}DWI} & \multicolumn{3}{c}{FLAIR--\textgreater{}T1} \\ \hline \hline
Method                    & PSNR $\uparrow$          & SSIM $\uparrow$          & RMSE $\downarrow$          & PSNR $\uparrow$           & SSIM $\uparrow$         & RMSE $\downarrow$         \\ \hline
None                      & 25.61         & 0.877         & 0.055         & 25.15         & 0.876        & 0.064        \\
+L                        & 25.66         & 0.876         & 0.055         & 25.21         & 0.878        & 0.062        \\
+H                        & 25.59         & 0.875         & 0.055         & 25.15         & 0.874        & 0.062        \\
+CrossAttn                 & 25.64         & 0.877         & 0.055         & 25.21         & 0.875        & 0.062        \\
+L, H                      & \textbf{25.68} & \textbf{0.877} & \textbf{0.054} & \textbf{25.31} & \textbf{0.878} & \textbf{0.061} \\ \hline
\end{tabular}
\end{small}
\end{table}

\section{Conclusion}
In this paper, we propose a new diffusion model architecture for medical image translation, which achieves generating target modalities with correct anatomical structures through the dynamic frequency balance module and the modality-centered knowledge-guided mechanism. Experimental results demonstrate our method's effectiveness in medical image translation and excellent performance in generating correct anatomical structures. Ablation experiments also demonstrate the effectiveness of our proposed dynamic frequency balance module and modality-centered knowledge-guided mechanism.

In summary, we propose a novel method for structure-accurate medical image translation, which aims to overcome the interference of modality absence on clinical diagnosis. The proposed method is expected to facilitate radiologists in making more precise diagnoses.

\bibliographystyle{unsrt}  
\bibliography{references}  
\end{document}